# Generative Artificial Intelligence Chatbots for Motivational Interviewing: A Scoping Review From System Design to Intervention Outcomes

Runze Hu[1], Jingqi Kong[1], Yang Yang[1], Yihang Yang[1], Jingyao Liu[1], Haizhou Tang[1], Shanghang Zhang[2], Zheng Liu[1#]

**Affiliations:**

[1] Department of Maternal and Child Health, School of Public Health, Peking University, Beijing, China;

[2] Institute for Visual Technology, School of Computer Science, Peking University, Beijing, China.

**Corresponding Author:**

Zheng Liu, Department of Maternal and Child Health, School of Public Health, Peking University, Beijing, China, liuzheng@bjmu.edu.cn, https://orcid.org/0000-0002-0405-2348.

## Abstract

**Background:** Motivational interviewing (MI) is a collaborative communication approach used to elicit autonomous motivation in health-related behavior change. Generative artificial intelligence (GenAI) provides new opportunities to deliver MI through conversational systems, but evidence on how these systems are designed, assessed, and translated into interventions remains fragmented.

**Objective:** This scoping review aimed to characterize the current evidence on GenAI-MI chatbots across system design, safety measures, MI quality, user perceptions, and intervention outcomes.

**Methods:** We conducted a scoping review in accordance with PRISMA-ScR. Studies published or publicly available from January 1, 2015 to June 2, 2026 were identified through PubMed, Web of Science, Scopus, PsycINFO, DBLP, IEEE Xplore, ACM Digital Library, arXiv, and ACL Anthology. Eligible studies used generative AI to generate MI-related chatbot responses or counselor utterances. Data were extracted using a predefined extraction framework and synthesized descriptively.

**Results:** Forty-seven reports comprising 48 studies were included. Twenty studies (41.7%) focused on system design without direct participant use, whereas 28 (58.3%) involved direct interaction with a GenAI-MI chatbot. Most systems were text based and disembodied, and 23 of 48 studies (47.9%) incorporated dynamic adaptation.

Safety measures were unevenly reported, with privacy and data protection (20/48, 41.7%) and safety-oriented content generation (17/48, 35.4%) more commonly described than automated (7/48, 14.6%) or human (6/48, 12.5%) risk monitoring. Among studies involving direct participant use, 21 of 28 (75.0%) reported informed consent or user education. Thirty of 48 studies (62.5%) assessed MI quality using observer- or client-based evaluations. Existing observer and client evaluations generally suggested that GenAI-MI chatbots could produce MI-consistent interactions. User perceptions were generally favorable, particularly for empathy, usability, helpfulness, and intention to use, although measurement approaches were heterogeneous. Eighteen of 48 studies (37.5%) reported intervention outcomes, including applications in physical activity, smoking cessation, and alcohol or substance use. Most intervention studies involved a single session, and only 3 of 18 (16.7%) evaluated repeated use over 10 days to 4 weeks. Positive findings were reported more consistently for short-term motivation outcomes than for sustained behavioral or functional change.

**Conclusions:** Current evidence suggests that GenAI-MI chatbots can deliver MI-consistent interactions that are generally perceived favorably by users, but evidence supporting sustained behavioral or functional change remains limited. Future research should strengthen runtime safety monitoring, standardize MI quality assessment, and use longer-term comparative designs with behavioral and functional outcomes to determine whether short-term motivational changes could translate into meaningful intervention effects.

# Introduction

Physical and mental health interventions often require ongoing communication to understand individuals' health status, willingness to change, and barriers to change, and to provide appropriate support accordingly. However, traditional counseling is constrained by factors such as limited professional capacity, service costs, and geographic accessibility, making sustained and individualized support difficult to provide at scale[1]. Chatbots and conversational agents have therefore increasingly been used to support health services by delivering health information, psychological support, and behavioral interventions through interactions with users[2]. Meanwhile, many early chatbots relied on predefined rules, keyword matching, or structured dialogue flows, with relatively limited ability to understand users' open-ended expressions and generate flexible responses based on the conversational context[3].

With the development of generative artificial intelligence (GenAI), particularly large language models (LLMs),

chatbots for psychological counseling and health interventions have gained greater capacity to handle open-ended user input. GenAI can generate natural language responses based on user input, system prompts, and external information, thereby offering greater potential to accommodate individualized user expressions. Previous reviews have summarized the effects of GenAI-based chatbots in interventions targeting diet, physical activity, and smoking cessation, with some evidence of effectiveness for smoking cessation among adults[4].

Motivational interviewing (MI) is a client-centered communication approach focused on eliciting motivation for change. It emphasizes helping individuals explore and resolve ambivalence about change through collaboration, acceptance, empathy, and support for autonomy[5]. Unlike general health education or advice giving, MI requires counselors to continuously identify clients' change talk, sustain talk, emotional responses, and motivational states during the conversation, and to select appropriate responses accordingly, such as open-ended questions, affirmations, reflective listening, summaries, information provision, and action planning. Thus, MI is not simply a form of natural language communication but a counseling approach with a clear theoretical foundation, specific technical requirements, and established standards for process quality. Coding systems such as the Motivational Interviewing Treatment Integrity (MITI)[6] and Motivational Interviewing Skill Code (MISC)[7] provide operationalized frameworks for evaluating the quality of MI delivery, enabling researchers to assess counseling processes in terms of MI adherence, reflection-to-question ratios, change talk, and sustain talk. These characteristics make MI dependent on flexible natural language interaction while also being guided by relatively well-defined conversational principles and evaluation criteria, providing a foundation for the use of GenAI in MI conversations. In this review, we use the term GenAI-MI chatbot to refer to chatbots that use generative AI to generate MI-related responses.

A substantial body of research on GenAI-MI chatbots has emerged, covering a wide range of topics. For example, some studies have focused on how generative AI can be used to generate dialogue that more closely adheres to MI principles[8], whereas others have explored the use of GenAI-MI chatbots for smoking cessation[9], physical activity promotion[10], and other health behavior interventions. However, this body of evidence has not yet been systematically synthesized. Although a previous review examined AI systems delivering MI[11], most of the included studies used rule-based rather than generative approaches. Moreover, the review was limited to studies that implemented health interventions, thereby overlooking research conducted during the early stages of GenAI-MI chatbot design.

This scoping review aims to systematically synthesize research on generative AI–based MI chatbots and to characterize the existing evidence across five domains: system design, safety measures, MI quality evaluation, user perceptions, and intervention outcomes. First, we characterize the design and evaluation landscape of the field, including how these systems are designed, what safety measures are used, and how MI quality and user perceptions are assessed. Second, we examine the intervention evidence, focusing on whether favorable findings in MI quality and proximal motivational outcomes extend to behavioral or functional outcomes. Specifically, this review seeks to address the following five research questions:

RQ1: What are the design characteristics of GenAI-MI chatbots?

RQ2: What safety measures are adopted in GenAI-MI chatbots?

RQ3: What methods are used to assess the MI quality of GenAI-MI chatbots, and what do these evaluations indicate about MI quality?

RQ4: How are users' perceptions of GenAI-MI chatbots assessed and what perceptions have been reported?

RQ5: What intervention outcomes have been reported for GenAI-MI chatbots?

# Methods

This review was conducted in accordance with the Preferred Reporting Items for Systematic Reviews and Meta-Analyses Extension for Scoping Reviews (PRISMA-ScR) Checklist[12] (Multimedia Appendix 1). It systematically synthesized research at the intersection of GenAI and MI. The review protocol was registered on June 1, 2026, and is publicly available at https://osf.io/xzw4g/overview.

## Search Strategy

We systematically searched for studies in English published or publicly available between January 1, 2015, and June 2, 2026. The search period began in 2015 because the application of natural language processing (NLP) and artificial intelligence technologies in digital health and psychological interventions has increased substantially since that time, allowing the search to more closely capture recent developments in automated MI dialogue systems. Multiple reports of the same study were linked and treated as a single study, with the most complete report used for data extraction. Later companion or updated reports were used only for studies already identified by the search. The databases searched were PubMed, Web of Science, Scopus, PsycINFO, DBLP, IEEE Xplore, ACM Digital Library, arXiv, and ACL Anthology. The search strategy was developed around two core concepts. The first comprised MI-related terms, including "motivational interviewing". The second comprised terms related to automated conversational systems, including "chatbot*", "conversational agent*", "dialogue system*", "dialog system*", "large language model*", "LLM", and "LLMs". Search fields and syntax were adapted to the requirements of each database. For PubMed, Medical Subject Headings (MeSH) terms — "Motivational Interviewing" [MeSH Terms], "Large Language Models" [MeSH Terms], and "Generative Artificial Intelligence" [MeSH Terms] — were added to the free-text Title/Abstract search to improve search sensitivity. Web of Science, Scopus, PsycINFO, IEEE Xplore, and ACM Digital Library were searched primarily within title, abstract, and keyword fields, whereas DBLP, arXiv, and ACL Anthology were searched using title and abstract fields. The complete search strategies are provided (Multimedia Appendix 2).

## Inclusion and Exclusion Criteria

Inclusion criteria: (1) Original studies published in English in which motivational interviewing (MI) was a central component and MI concepts were explicitly incorporated into the study objectives and system design; (2) studies that used generative artificial intelligence to develop a chatbot or a related module for response generation. Generative AI included large language models or other neural generative language models capable of generating natural language responses based on user input, task instructions, or other contextual information; (3) studies that investigated either a complete GenAI-MI chatbot or a specific response-generation module intended for use within a GenAI-MI chatbot.

Exclusion criteria: (1) Studies in which MI was not explicitly a central component; (2) studies that did not use generative AI and relied only on rule-based scripts, fixed response libraries, decision trees, or conventional machine-learning classifiers; (3) studies that did not involve chatbot response generation or MI counselor utterance generation and instead focused solely on tasks such as automated MI coding, sentiment analysis, or other general natural language processing tasks; (4) reviews, conference abstracts, study protocols, opinion articles, book chapters, studies for which the full text was unavailable, non-English full-text publications, and duplicate publications.

## Study Selection

All retrieved records were imported into EndNote for reference management and deduplication. After duplicates were removed, two reviewers (RZH and JQK) independently screened the titles and abstracts according to the predefined inclusion and exclusion criteria. Full texts were obtained for records for which eligibility could not be determined based on the title and abstract alone.

During full-text screening, the same two reviewers independently assessed each article for eligibility. Any disagreements were first resolved through discussion between the two reviewers. If consensus could not be reached, a third reviewer (ZL) reviewed the article and made the final decision.

## Data Extraction

A prespecified data extraction form was used to extract information from the included studies (Multimedia Appendix 3). For all included studies, we extracted general study characteristics, system design features of the GenAI-MI chatbot, safety measures, and methods used to evaluate MI quality. System design characteristics included interaction modality, type of embodiment, interaction language, and dynamic adaptation. For MI quality evaluation, we extracted the evaluation metrics and the corresponding findings.

For studies in which participants directly interacted with a GenAI-MI chatbot, we additionally extracted the study objectives, study design, study population, sample size, study conditions, and the number or duration of interactions, as summarized in Supplementary Table 3. For studies assessing user perceptions, we extracted the scales used, items developed by the researchers, and findings from qualitative interviews. For studies reporting intervention outcomes, we extracted motivation outcomes and behavioral or functional outcomes, together with their measurement time points, scale ranges, and outcome data available for quantitative calculation.

Data extraction was conducted independently by two reviewers (RZH and JQK). Any discrepancies were first resolved through discussion between the two reviewers. If consensus could not be reached, a third reviewer (ZL) reviewed the data and made the final decision.

## Data Synthesis

Because the included studies showed substantial heterogeneity in system design, evaluation methods, study design, and outcome measures, we primarily used descriptive statistics and narrative synthesis and did not pool effect sizes across studies. Based on the objectives and actual content of each study, studies were classified into three study types[13]. Design studies referred to studies involving the development, optimization, or technical validation of GenAI-MI chatbots, key generation modules, or dialogue strategies. Feasibility studies referred to studies in which participants directly interacted with a GenAI-MI chatbot and evaluated its user experience, acceptability, or implementation feasibility. Evaluation studies referred to studies examining motivational, behavioral, or functional outcomes following the use of GenAI-MI chatbots. These three study types were not mutually exclusive, and a single study could be classified into more than one type.

For all included studies, we synthesized the evidence across three domains: system design, safety measures, and MI quality evaluation. System design characteristics were summarized using narrative synthesis. Safety measures were categorized according to the measures actually reported in the included studies[14]. MI quality evaluation methods were synthesized according to the type of assessment used, including natural language generation metrics, MI-specific observer evaluations, and MI-specific client evaluations[15].

For studies in which participants directly interacted with a GenAI-MI chatbot, that is, studies classified as feasibility, evaluation, or both, we conducted an additional descriptive synthesis. Because the labels used for intervention and control conditions in the original studies varied according to their respective research questions, we reclassified study conditions using a standardized framework. Conditions in which GenAI was primarily responsible for response generation and MI served as the principal intervention approach were classified as GenAI-MI conditions, whereas all other conditions used for comparison were classified as comparator conditions.

For studies involving direct participant use that reported users' perception outcomes, we conducted construct mapping of participants' perceptions of GenAI-MI chatbots. Because similar user perceptions were assessed using different measures across studies, outcomes were not classified solely according to the terminology used in the original studies. Instead, classification was primarily based on the theoretical definition of each measure and the content it was intended to assess. Items developed by the researchers and qualitative interview themes were classified according to their actual evaluative content. Two reviewers (RZH and JQK) independently coded the constructs, grouping measures with the same or highly similar theoretical meaning into a common construct. Disagreements were resolved through discussion. This approach was informed by previous reviews of conversational agents that mapped heterogeneous evaluation measures onto common constructs[3, 16]. A single study could contribute to multiple constructs but was counted only once within each construct. The resulting constructs and their operational definitions are reported in the Results section.

For evaluation studies, we focused on motivation outcomes and behavioral or functional outcomes. Motivation outcomes were operationally defined as proximal psychological indicators that directly reflected individuals' willingness, readiness, or confidence to change a target behavior, including importance, readiness, confidence, self-efficacy, and intention to change. Behavioral or functional outcomes were defined as indicators reflecting whether the target behavior was actually performed or whether behavioral levels or functional status changed. Because relatively few studies reported behavioral or functional outcomes, and because these outcomes varied substantially in their definitions, measurement units, and reporting formats, they were synthesized narratively.

For motivation outcomes, descriptive quantitative analyses were limited to measures for which both baseline and post-intervention means were reported and the theoretical scale range was clearly defined. Range-normalized change scores were calculated relative to the theoretical range of each scale as follows:

$$Change = \frac{M(post) - M(pre)}{U - L} \times 100\%$$

*Change* represents the range-normalized change score; $M_{post}$ represents the mean after the intervention; $M_{pre}$

represents the baseline mean. $U$ and $L$ represent the theoretical maximum and minimum values of the scale. Positive values indicate an increase in motivation outcomes after the intervention. For single-arm studies, pre-post changes were calculated within the study condition. For studies with multiple conditions, changes were calculated separately for each condition. When a study condition included multiple motivation outcomes that met the calculation criteria, their range-normalized change scores were averaged using an unweighted arithmetic mean to reduce duplicate representation of the same condition due to the reporting of multiple related outcomes[17]. This mean was used solely as a condition-level descriptive summary and was not considered a psychometrically validated measure. Finally, changes in the GenAI-MI chatbot conditions and their corresponding comparator conditions were presented in a descriptive dot plot. Consistent with the evidence-mapping purpose of this scoping review, we did not undertake a formal risk of bias or critical appraisal of the included studies. Accordingly, intervention-related findings were synthesized descriptively and were not used to draw conclusions regarding intervention efficacy. The overall framework for data extraction and synthesis is presented below (Table 1).

**Table 1.** Overview of Data Extraction and Synthesis

| Study subset | Data extracted | Data Synthesis | Research question |
|---|---|---|---|
| All included studies | System design characteristics | Narrative synthesis | RQ1 |
| All included studies | Safety measures | Narrative synthesis | RQ2 |
| All included studies | Methods for evaluating MI quality and corresponding evaluation findings | Narrative synthesis | RQ3 |
| Studies involving direct participant use that reported user perception outcomes | User perceptions, scales, items developed by the researchers, and qualitative interview findings | Construct mapping based on the theoretical definitions and actual content of the measures; study counts for each construct; narrative synthesis | RQ4 |
| Evaluation studies | Motivation outcomes; behavioral or functional outcomes | Range-normalized changes were calculated for motivation outcomes when baseline and follow-up means and the theoretical scale range were available; outcomes that could not be calculated were narratively synthesized. Behavioral or functional outcomes were narratively synthesized. | RQ5 |

# Results

## Study Characteristics

The search identified 723 records, of which 265 were duplicates and 372 were excluded during title and abstract screening. The full texts of the remaining 86 articles were reviewed, resulting in the inclusion of 47 articles comprising 48 independent studies. Figure 1 summarizes the study identification and screening process.

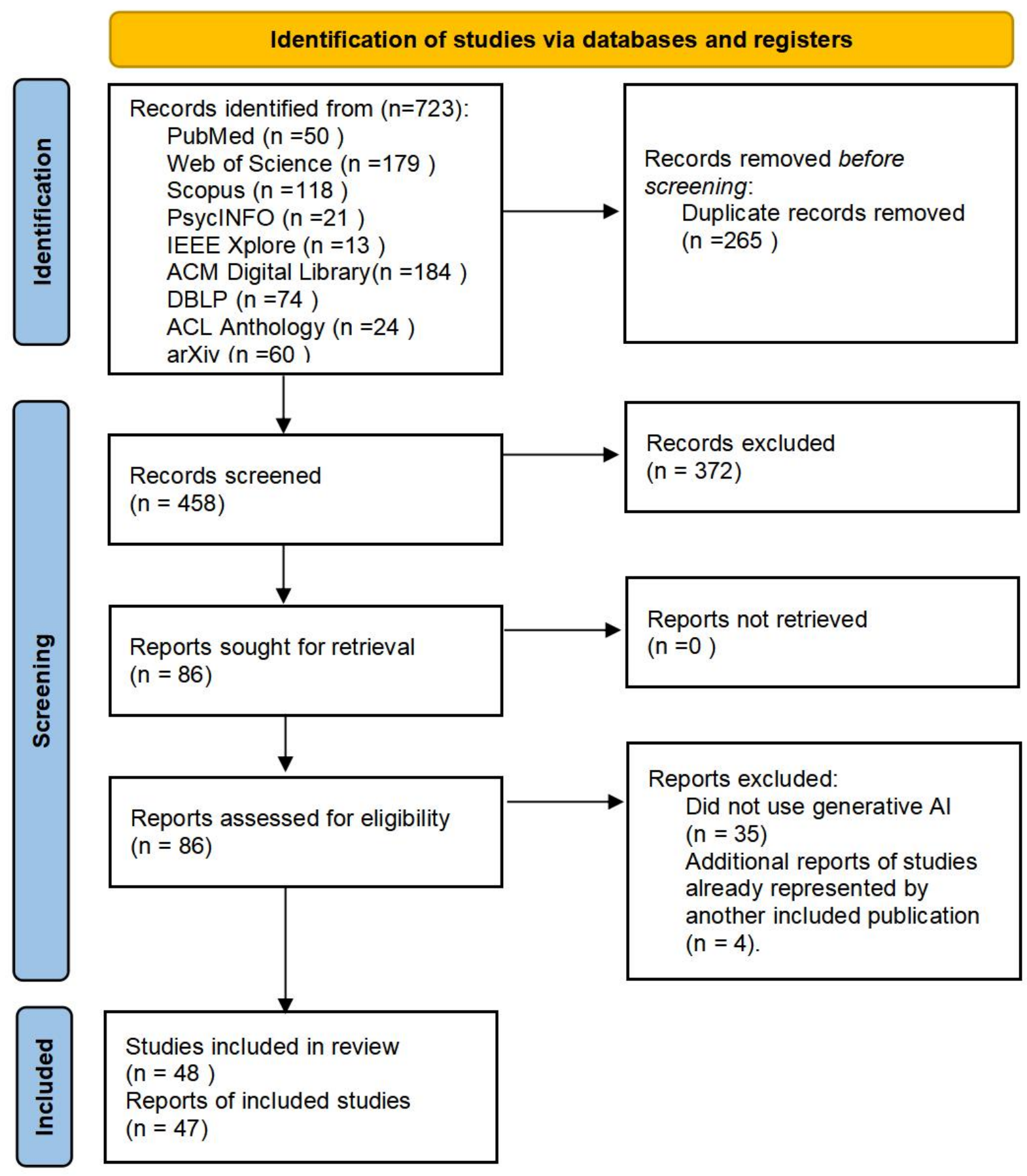

Figure 1. Flow diagram of study identification, screening, and inclusion

All included studies were published in 2023 or later, and no studies meeting the eligibility criteria were identified between 2015 and 2022. By publication year, 3 studies (6.3%) were published in 2023, 9 (18.8%) in 2024, 22 (45.8%) in 2025, and 14 (29.2%) in 2026. Of the 48 studies, 40 (83.3%) were classified as Design, 27 (56.3%) as Feasibility, and 18 (37.5%) as Evaluation. Twenty studies (41.7%) were classified as design only, without direct participant use of a GenAI-MI chatbot. In the remaining 28 studies (58.3%), participants directly interacted with a GenAI-MI chatbot as part of feasibility, evaluation, or both. Among the 28 studies involving direct participant use of a GenAI-MI chatbot, study designs were diverse: 7 were randomized controlled trials, 3 were non-randomized between-subject comparative studies, 5 were within-subject comparative studies, 2 used both between- and within-subject comparisons, 4 were uncontrolled pre-post studies, and 7 were uncontrolled post-only studies. Sixteen studies (57.1%) had sample sizes below 50 participants, 4 (14.3%) included 50-100 participants, and 8 (28.6%) included more than 100 participants. Most studies involved a single interaction with the chatbot (21/28, 75.0%), while 4 (14.3%) involved interactions more than once or comparative exposures and 3 (10.7%) evaluated repeated use over periods ranging from 10 days to 4 weeks (Table 2).

**Table 2**. Characteristics of Included Studies

| Characteristic | Category | Number of studies | Percentage |
|---|---|---|---|
| Study type | Design only | 20 | 41.7% |
| | Feasibility only | 1 | 2.1% |
| | Evaluation only | 1 | 2.1% |
| | Design and feasibility | 9 | 18.8% |
| | Feasibility and evaluation | 6 | 12.5% |
| | Design and evaluation | 0 | 0.0% |
| | Design, feasibility, and evaluation | 11 | 22.9% |
| User involvement | Direct participant use [a] | 28 | 58.3% |
| | No direct participant use | 20 | 41.7% |
| Publication year | 2023 | 3 | 6.3% |
| | 2024 | 9 | 18.8% |
| | 2025 | 22 | 45.8% |
| | 2026 | 14 | 29.2% |
| Study design (Studies involving direct participant interaction, n=28) | Randomized controlled trial | 7 | 25.0% |
| | Non-randomized between-subject comparative study | 3 | 10.7% |
| | Within-subject comparative study | 5 | 17.9% |
| | Mixed between- and within-subject comparative study | 2 | 7.1% |
| | Uncontrolled pre-post study | 4 | 14.3% |
| | Uncontrolled post-only study | 7 | 25.0% |
| Study sample size (Studies involving direct participant interaction, n=28) | <50 | 16 | 57.1% |
| | 50 – 100 | 4 | 14.3% |
| | >100 | 8 | 28.6% |
| Interaction schedule (Studies involving direct participant interaction, n=28) | Single interaction | 21 | 75.0% |
| | 2 – 4 interactions/comparative exposures | 4 | 14.3% |
| | Repeated use over 10 days – 4 weeks | 3 | 10.7% |

a: Studies with direct participant use were those classified as feasibility, evaluation, or both.

## System Design Characteristics

Among the 48 included studies, text remained the predominant interaction modality. Thirty-nine studies (81.3%) used text-only interaction, whereas the remaining 9 studies (18.8%) incorporated speech or visual information in the input or output. When classified by modality combination, 1 study (2.1%) supported both text and speech interaction[18], 2 studies (4.2%) combined text interaction with visual information such as charts[19, 20], and 5 studies (10.4%) combined speech and visual information, including facial expressions, virtual avatars, physical robots, or immersive virtual environments[21-25]. One additional study (2.1%) integrated text, speech, and visual information[26]. Overall, current GenAI-MI chatbots remain predominantly text-based, although some studies have expanded interaction through speech, visual presentation, and nonverbal signals.

Most systems were disembodied, accounting for 42 studies (87.5%). Four studies (8.3%) presented the chatbot through a virtual avatar or virtual environment[23-26], 1 study (2.1%) used a physical robot[22], and another study (2.1%) included both virtual and physical embodiments[21]. Thus, although conventional disembodied interfaces remain dominant, a small number of studies have begun to explore virtual agents, social robots, and immersive environments as alternative forms of delivering GenAI-based MI.

English was the predominant interaction language, used in 40 studies (83.3%). French was used in 3 studies (6.3%)[21, 25, 27], while German[28], Chinese[29], Korean[30], and Japanese[31] were each used in 1 study (2.1%). One additional study (2.1%) used both English and Spanish[32].

For dynamic adaptation, 25 studies (52.1%) did not report an explicit mechanism for adapting dialogue strategies or flow during interaction. These systems mainly relied on fixed prompts and predefined instructions. The remaining 23 studies (47.9%) incorporated dynamic adaptation, whereby information obtained during the interaction influenced subsequent system behavior. Common dynamic adaptation approaches included inferring dialogue or motivational states and selecting corresponding MI strategies[8, 30, 31, 33, 34], retrieving relevant knowledge, demonstrations, or strategy information based on the current conversation[32, 35, 36], incorporating user information, longitudinal interaction history, or data from wearable devices to support personalized responses[19, 20, 32, 37], and using nonverbal information such as facial expressions or other multimodal signals to adapt system behavior[21, 22]. Some systems also dynamically adjusted predefined dialogue pathways according to the current conversational state[10, 18, 24, 25, 27, 38]. Overall, dynamic adaptation in GenAI-MI chatbots extended beyond fixed prompting to include MI strategy selection, knowledge retrieval, modeling of user states, multimodal information, and adjustment of dialogue flow (Table 3).

**Table 3**. System Design Characteristics of Included Studies

| Characteristic | Category | Number of studies | Percentage | Studies |
|---|---|---|---|---|
| Interaction modality | Text | 39 | 81.3% | All except the 9 studies below |
| | Text and speech | 1 | 2.1% | [18] |
| | Text and visual | 2 | 4.2% | [19, 20] |
| | Speech and visual | 5 | 10.4% | [21-25] |
| | Text, speech, and | 1 | 2.1% | [26] |

| Characteristic | Category | Number of studies | Percentage | Studies |
| --- | --- | --- | --- | --- |
| | visual | | | |
| Type of embodiment | Disembodied | 42 | 87.5% | All except the 6 studies below |
| | Virtual representation | 4 | 8.3% | [23-26] |
| | Physical representation | 1 | 2.1% | [22] |
| | Combined | 1 | 2.1% | [21] |
| Interaction language | English only | 40 | 83.3% | All except the 8 studies below |
| | French | 3 | 6.3% | [21, 25, 27] |
| | German | 1 | 2.1% | [28] |
| | Chinese | 1 | 2.1% | [29] |
| | Korean | 1 | 2.1% | [30] |
| | Japanese | 1 | 2.1% | [31] |
| | English and Spanish | 1 | 2.1% | [32] |
| Dynamic adaptation | No | 25 | 52.1% | [9, 23, 26, 29, 39-59] |
| | Yes | 23 | 47.9% | [8, 18-22, 24, 25, 27, 28, 30-38, 60, 61], Sun et al. Study 1[10], Sun et al. Study 2[10] |

## Safety Measures

Safety measures were classified into five constructs that were not mutually exclusive: safety-oriented content generation, automated risk monitoring, human risk monitoring, privacy and data protection, and informed consent and user education. Among the 48 studies, 17 (35.4%) implemented safety-oriented content generation, 7 (14.6%) automated risk monitoring, 6 (12.5%) human risk monitoring, and 20 (41.7%) privacy and data protection measures. Among the 28 studies involving direct participant use, 21 (75.0%) reported informed consent or user education.

Safety-oriented content generation referred to measures incorporated into prompts, knowledge sources, or generation workflows to reduce the production of harmful, inaccurate, inappropriate and biased responses[10, 19, 21, 24-26, 28, 32, 37-39, 43, 44, 48, 54, 58, 61]. Some studies constrained model generation using curated or validated knowledge sources to reduce unsupported or inaccurate information[32, 48]. Others incorporated explicit safety instructions into system prompts, such as prohibiting diagnosis, prescribing, dangerous advice, or discussion beyond predefined support boundaries[21, 44]. Some systems further restricted the range of allowable responses through structured dialogue workflows or predefined expert-generated content, as reported in Study 2 by Sun et al.[10] and in another study[24].

Automated risk monitoring referred to mechanisms that automatically detected potential risks in user inputs or model responses and took corresponding actions to prevent unsafe content from reaching users[19, 32, 37-39, 43, 61]. Some studies automatically screened conversational content for predefined risks, including self-harm, inappropriate disclosure of personal information, or unsupported speculative content[39]. Others used an independent LLM or reviewer agent to evaluate candidate responses for safety before they were delivered and to request revision when necessary[61]. Some systems used classifiers to identify harmful or otherwise inappropriate outputs and automatically rewrite them before delivery[19, 38]. In addition, some systems repeatedly regenerated responses when an automated moderator determined that the generated content did not meet predefined safety requirements[43]. Automated monitoring was also used to flag potentially risky interactions for subsequent review or escalation[32].

Human risk monitoring referred to review or supervision by researchers or professionals who could intervene, escalate concerns, or provide support when a potential risk was identified[20, 23-25, 32, 42]. In some studies, researchers, operators, or mental health professionals supervised interactions and were available to intervene when abnormal responses or potential emergencies occurred[20, 23, 24, 32]. Other studies incorporated post-interaction manual review, with identified risks followed by predefined escalation procedures[42].

Privacy and data protection referred to measures designed to minimize the collection or disclosure of sensitive information and to protect data during processing, transmission and storage[18-20, 23, 27, 31, 32, 34, 37, 40, 42-46, 50, 54, 56, 57, 59]. Some studies used storage environments compliant with health data security requirements together with data encryption[32]. Other studies implemented encrypted transmission, authentication, and encrypted storage[19], or used access controls and de-identification to protect user data[44]. In addition, one study used a locally deployed model to remove privacy-sensitive information from conversations[37], while another stored users’ historical data on their own devices[18].

Informed consent and user education primarily included obtaining informed consent before study participation, providing instructions on system use, and informing users about relevant risks or limitations of system use[9, 10, 18-21, 23-25, 28, 31, 33, 38, 41, 43-47, 58, 59]. Several studies obtained written or electronic informed consent before participants used the chatbot[9, 24, 44], including Study 1 by Sun et al.[10]. Some studies additionally provided instructions on system use. For example, participants were introduced to how the system operated before formal use[18], while another study informed participants about the use of LLMs and associated potential risks during the study[31].

## MI Quality

Among the 48 included studies, 5 (10.4%) used natural language generation metrics to assess the quality of MI dialogues generated by GenAI-MI chatbots. Common metrics included BLEU and ROUGE for lexical overlap, METEOR and BERTScore for lexical or semantic similarity, MAUVE for distributional similarity, and metrics such as perplexity and Distinct for fluency and diversity. One study found that fine-tuning improved scores on natural language generation metrics, indicating greater similarity to reference responses[29]. Similarly, a benchmark study found better performance on natural language generation metrics with strategy-aligned generation than with standard prompting[34]. However, other studies found that these metrics showed limited ability to distinguish MI quality or were inconsistent with expert assessments. One study found that responses generated by a GenAI-MI chatbot showed relatively low linguistic similarity to human therapist responses despite high consistency in meaning[39]. Another study found that natural language generation metrics did not effectively distinguish therapist performance levels validated by MI experts[50]. Similarly, differences across LLMs were limited on natural language generation metrics but became more apparent when indicators specific to MI were incorporated[60].

Beyond these general language generation metrics, 30 studies used observer-based or client-reported approaches to assess dialogue quality. Observer evaluation was used in 24 studies (50.0% of all included studies), whereas 9 studies (18.8%) assessed MI quality from the client perspective. Three studies used both observer evaluation and client evaluation (Table 4). Observer evaluations examined whether chatbot responses adhered to MI principles and demonstrated MI behaviors, such as reflections, open questions and elicitation of change talk. Observer evaluations generally indicated that GenAI-MI chatbots were capable of delivering responses with high MI quality. For example, one benchmark study evaluating 10 LLMs with MITI found high global MI scores across models, with some behavioral indicators reaching or exceeding those observed in real clinical conversations[40]. Similarly, a comparison between a GPT-4 virtual counselor and human counselors reported MI adherence rates of 94% and 96%[26].

Automated coding of MI quality has also begun to emerge. Among the 24 studies, 13 (54.2%) relied on human evaluators, 5 (20.8%) used fully automated assessment, and 6 (25.0%) combined human and automated assessment. Human evaluators included MI coders, clinicians, counseling experts, and trained research staff, whereas automated evaluators included LLMs and classification models trained to identify MI behaviors. Validation against human coding showed moderate agreement in several studies. For example, one study used GPT-4o to automatically code counseling dialogues according to MISC and reported a Fleiss κ of 0.68 compared with four human evaluators[43]. Another study found that GPT-4, when used as a zero-shot classifier of MI reflections, achieved a Cohen κ of 0.66 with human reviewers[57]. These findings indicate that automated MI coding may support more scalable dialogue assessment, although continued validation against human annotations remains necessary.

From the client perspective, 9 studies used client measures to assess clients' perceptions of the MI process. These measures assessed aspects of MI that are directly experienced by clients. The Client Evaluation of Motivational Interviewing (CEMI)[15] was the most commonly used measure[21, 25, 26, 28, 33, 42, 46]. Two other studies used adapted measures of perceived MI adherence[10]. Client findings indicated generally high perceived MI quality. One study found that CEMI scores were significantly above the predefined treatment threshold[26]. Another reported increases in both relational and technical CEMI subscales after iterative chatbot refinement, although only the improvement in the relational subscale was statistically significant[46]. In a feasibility study of 98 emergency department patients, the relational subscale was rated high (mean=85.3/100), whereas the technical subscale was rated moderate (mean=68.4/100)[42].

**Table 4**. Methods for Evaluating MI Quality in GenAI-MI Chatbots

| Dimension | Category | Number of studies | Percentage | Studies |
|---|---|---|---|---|
| Assessment approach | Natural language generation metrics | 5 | 10.4% | [29, 34, 39, 50, 60]. |
| | Observer-based | 24 | 50.0% | [8, 9, 20, 26, 29-34, 36, 37, 40, 43, 45-47, 51, 54-57, 60, 61] |
| | Client-reported | 9 | 18.8% | [21, 25, 26, 28, 33, 42, 46], Sun et al. Study 1[10], Sun et al. Study 2[10] |
| Observer (Studies involving observer evaluation, n=24) | Human | 13 | 54.2% | [20, 26, 29-32, 34, 40, 45, 46, 54-56] |
| | Automated | 5 | 20.8% | [33, 43, 47, 51, 61] |
| | Mixed | 6 | 25.0% | [8, 9, 36, 37, 57, 60] |

## User Perceptions

Of the 28 studies involving direct participant use of a GenAI-MI chatbot, 27 reported participants' perceptions of the chatbot or interaction and were included in the users' perception synthesis. These reported perceptions were mapped into 13 user perception constructs, such as perceived empathy, helpfulness, usability, and therapeutic alliance (Table 5). Empathy was the most frequently evaluated construct, assessed in 13 studies (48.1%), followed by helpfulness in 12 studies (44.4%) and usability in 11 studies (40.7%). Therapeutic alliance was assessed in 9 studies (33.3%), and interaction quality in 8 studies (29.6%). Intention to use and satisfaction were each assessed in 7 studies (25.9%), while engagement was assessed in 6 studies (22.2%). Rapport, disclosure, and personalization were each evaluated in 4 studies (14.8%). Social presence and trust were the least frequently evaluated constructs, each appearing in only 2 studies (7.4%).

Measurement approaches for user perceptions were heterogeneous across studies. Standardized instruments and adapted versions were used, including the Consultation and Relational Empathy (CARE) Measure for empathy[43], the System Usability Scale (SUS)[42, 46] and Bot Usability Scale (BUS-15)[38] for usability, and the Working Alliance Inventory (WAI)[33] or Session Rating Scale (SRS) for therapeutic alliance[35]. Other studies used specific Likert questionnaires, single-item ratings, or qualitative interviews to assess constructs such as helpfulness, interaction quality, intention to use, disclosure, personalization, social presence, and trust.

Despite this methodological heterogeneity, user perceptions were generally favorable across most constructs. Empathy scores included 42/50 on the CARE Measure[43] and 4.5/5 in another study[20]. Usability was consistently rated positively, with SUS scores of approximately 81-85/100[42, 46] and 88.9% of participants in one study reporting that the system was easy to use[23]. Helpfulness was also generally rated highly, with 85% of participants in one study considering the system beneficial[23]. Therapeutic alliance and interaction quality varied across systems, although most ratings remained moderate to high. For example, one study reported conversational quality ratings above 4.5/5[18], while SRS scores were approximately 32/40 in another[35].

Intention to use, satisfaction, and engagement were similarly positive overall. Across several studies, approximately 72% – 90% of participants indicated willingness to continue using the systems[23, 42, 44], and one study reported that 92% of participants enjoyed the interaction[43]. Rapport, disclosure, and personalization were assessed less frequently but were also generally positive. For example, approximately 80% of participants in one study reported being able to open up to the virtual agent[23], and personalization ratings reached 4.75/5 in another[18]. Social presence and trust were examined in only a small number of studies, with quantitative ratings generally positive, although qualitative findings also identified concerns regarding data confidentiality and trust[26].

User perceptions also varied across GenAI-MI system configurations. In one study, usability scores across three chatbot configurations were 3.56, 3.61, and 3.82/5, with the highest rating observed when expert scripts were incorporated into the prompt[10]. Another study reported a higher WAI score for a system that dynamically selected MI strategies according to users' motivational states than for a basic MI system[33]. Differences in rapport were also observed across facial expression conditions, although real-time matching of facial expressions did not clearly outperform the condition without expression[21]. These findings indicate that user perceptions may vary according to strategy control and interaction design, even among systems that all incorporate GenAI and MI.

Overall, user perceptions of existing GenAI-MI chatbots were generally positive, particularly with respect to usability, helpfulness, empathy, and intention to use. However, user perceptions varied across system configurations, while trust and social presence remained particularly underexamined.

**Table 5**. User Perception Constructs and Operational Definitions for GenAI-MI Chatbots

| Construct | Operational Definition | **n** | Proportion of the 27 studies | Studies |
|---|---|---|---|---|
| Empathy | The extent to which users perceive that the chatbot understands their situation and emotions and responds with care, acceptance, and a nonjudgmental attitude. | 13 | 48.1% | [9, 19, 20, 22, 28, 31, 35, 43, 44, 47, 59], Sun et al. Study 1[10], Sun et al. Study 2[10] |
| Helpfulness | The extent to which users perceive that the chatbot provides helpful support or facilitates goal attainment. | 12 | 44.4% | [18-20, 23, 24, 31, 38, 43, 45, 47, 52, 58] |
| Usability | The extent to which users perceive the system as easy to learn, easy to use, easy to understand, and convenient to operate overall. | 11 | 40.7% | [18-20, 23, 26, 38, 42, 45, 46], Sun et al. Study 1[10], Sun et al. Study 2[10] |
| Therapeutic alliance | The extent to which users perceive a collaborative and supportive relationship with the chatbot in working toward shared goals. | 9 | 33.3% | [19, 20, 24, 28, 33, 35, 38], Sun et al. Study 1[10], Sun et al. Study 2[10] |
| Interaction quality | The extent to which users perceive the chatbot's language, response relevance, naturalness, and interaction fluency as appropriate. | 8 | 29.6% | [18, 20, 21, 24, 28, 31], Sun et al. Study 1[10], Sun et al. Study 2[10] |
| Intention to use | The extent to which users are willing to use the system again or continue using it in the future and recommend it to others. | 7 | 25.9% | [18, 22-24, 42, 44, 58] |
| Satisfaction | The extent to which users are satisfied with and like the system or interaction overall. | 7 | 25.9% | [18, 24, 43, 45, 53, 58, 59] |
| Engagement | The extent to which users are engaged, attentive, and involved during interactions with the chatbot. | 6 | 22.2% | [22, 28, 53, 58], Sun et al. Study 1[10], Sun et al. Study 2[10] |
| Rapport | The extent to which users perceive harmony, closeness, and interpersonal connection during interactions with the chatbot. | 4 | 14.8% | [21, 23-25] |
| Disclosure | The extent to which users are willing to disclose personal information, feelings, or conversation content. | 4 | 14.8% | [19, 20, 23, 42] |
| Personalizatio n | The extent to which users perceive that the chatbot tailors its responses or support to their individual needs, characteristics, preferences, or context. | 4 | 14.8% | [18-20, 44] |
| Social presence | The extent to which users perceive the chatbot as socially present or experience the interaction as involving a socially responsive conversational partner. | 2 | 7.4% | [21, 22] |
| Trust | The extent to which users perceive the chatbot and its responses as reliable, credible, and worthy of confidence. | 2 | 7.4% | [24, 26] |

Note: Percentages were calculated using the 27 studies included in the users' perception synthesis. Each study could contribute to multiple constructs but was counted only once within each construct. Sun et al.[10] included two independently coded studies.

## Intervention Outcomes

Among the 28 studies in which participants directly interacted with a GenAI-MI chatbot, 18 further reported intervention outcomes. The most common application areas were physical activity (n=5 studies from 4 reports)[10, 19, 33, 58] and smoking cessation (n=4)[9, 43, 44, 59], while 2 studies focused on reducing alcohol use[24, 25]. The remaining studies addressed dietary change[45], colorectal cancer screening[47], prosocial behavior[41], substance use in emergency department settings[42], daily physical and mental health and functioning[18], multiple health behaviors[28] and learning related to the Sustainable Development Goals[53]. Of these 18 studies, 17 reported motivation outcomes, 5 reported behavioral or functional outcomes, and 4 reported both types of outcomes. Fifteen of the 18 studies involved only a single intervention session. Only 3 studies repeated use over periods ranging from 10 days to 4 weeks[10, 18, 19]. Overall, existing intervention studies have primarily focused on short-term changes in motivation, whereas sustained use and behavioral or functional outcomes have been reported less frequently.

Figure 2 descriptively presents range-normalized changes in motivation outcomes for studies with sufficient data. Across the included GenAI-MI conditions, the calculated changes were positive, ranging from 2.9% to 16.0%. In studies with comparator conditions, the magnitude of these descriptive changes varied across study conditions. Because these values represent within-condition pre-post changes rather than between-group effect estimates, they should not be interpreted as evidence of comparative effectiveness. For example, in one study of prosocial motivation, the range-normalized changes were 8.2% in the GenAI-MI condition and 2.0% in a GPT-4 comparator without MI adaptation[41]. In another comparative study, the corresponding changes were 8.7% and 7.3%, and the original study reported no statistically significant between-group difference ($p=0.967$)[28]. In the cancer screening study[47], the mean range-normalized change was 9.2% in the GenAI-MI condition and 9.4% in one comparator condition, which received only a single personalized AI-generated screening message, indicating little numerical difference between the conditions. In a study aimed at increasing prosocial motivation, however, the short-term increase was not maintained at the 24-hour follow-up[41].

The magnitude of motivational change also varied across different configurations of GenAI-MI chatbots. In one comparison of two GenAI-MI configurations, the condition incorporating active listening showed a range-normalized change of 7.0%, compared with 2.9% without the strategy[58]. Another study compared two GenAI-MI chatbots and reported range-normalized changes of 14.7% for the dynamically adapted condition and 8.4% for the basic GenAI-MI condition[33]. However, the between-condition difference did not meet the significance threshold after correction for multiple comparisons[33]. Across three successive versions of a GenAI-MI chatbot in smoking cessation, range-normalized changes were 5.7%, 6.0%, and 8.0%[9]. These findings illustrate variation in motivation-related changes across system configurations but do not establish that more complex configurations were more effective.

Several studies that could not be included in Figure 2 because of insufficient continuous pre-post data nevertheless reported motivation outcomes. In a dietary intervention, the original study reported a significant increase in immediate intention to change diet ($p<0.01$), with a greater increase than in the comparator condition[45]. In a study of sustainable development goal learning, participants interacting with a GenAI-MI chatbot reported greater interest in learning than those receiving directive communication ($p=0.032$), as well as higher conversational engagement ($p<0.01$)[53]. Among emergency department patients with substance use, readiness to change increased in 24% of participants after a single GenAI-MI chatbot session, decreased in 13%, and remained unchanged in 63%[42].

Behavioral and functional outcomes were reported less frequently than motivation outcomes, and results varied across studies. In one smoking-cessation study, 70.5% of participants reported some reduction in smoking at 1 week, but the number of quit attempts did not change significantly and no significant differences were observed across chatbot versions[9]. In a 4-week physical activity study, the proportion of participants meeting the recommended 150 minutes per week increased from 31% to 74% in the GenAI-MI condition and from 41% to 71% in the comparator condition[19]. However, objective measures, including daily step counts, showed no significant between-group differences. In another 10-day physical activity study, daily step counts generally increased in both GenAI-MI conditions but decreased in the rule-based chatbot condition[10]. However, the observation period was short, making it unclear whether these differences would be sustained.

Evidence for behavioral change in other application areas remains inconclusive. In the prosocial-behavior study, the GenAI-MI condition did not significantly increase participants’ effort toward prosocial goals in an experimental task[41]. In a 14-day study of 20 participants, the number of daily functioning domains classified as having or possibly having functional problems was significantly lower at the end of the intervention than at baseline ($p<0.01$)[18]. Four participants continued to a 24-week follow-up, and none reported further difficulties in daily functioning. Because the long-term follow-up included only 4 participants, these findings remain exploratory. Overall, positive short-term changes were reported more frequently for motivation outcomes, whereas sustained behavioral and functional outcomes were examined less often and showed greater variability.

**Figure 2.** Descriptive dot plot of range-normalized changes in motivation outcomes across GenAI-MI and comparator conditions

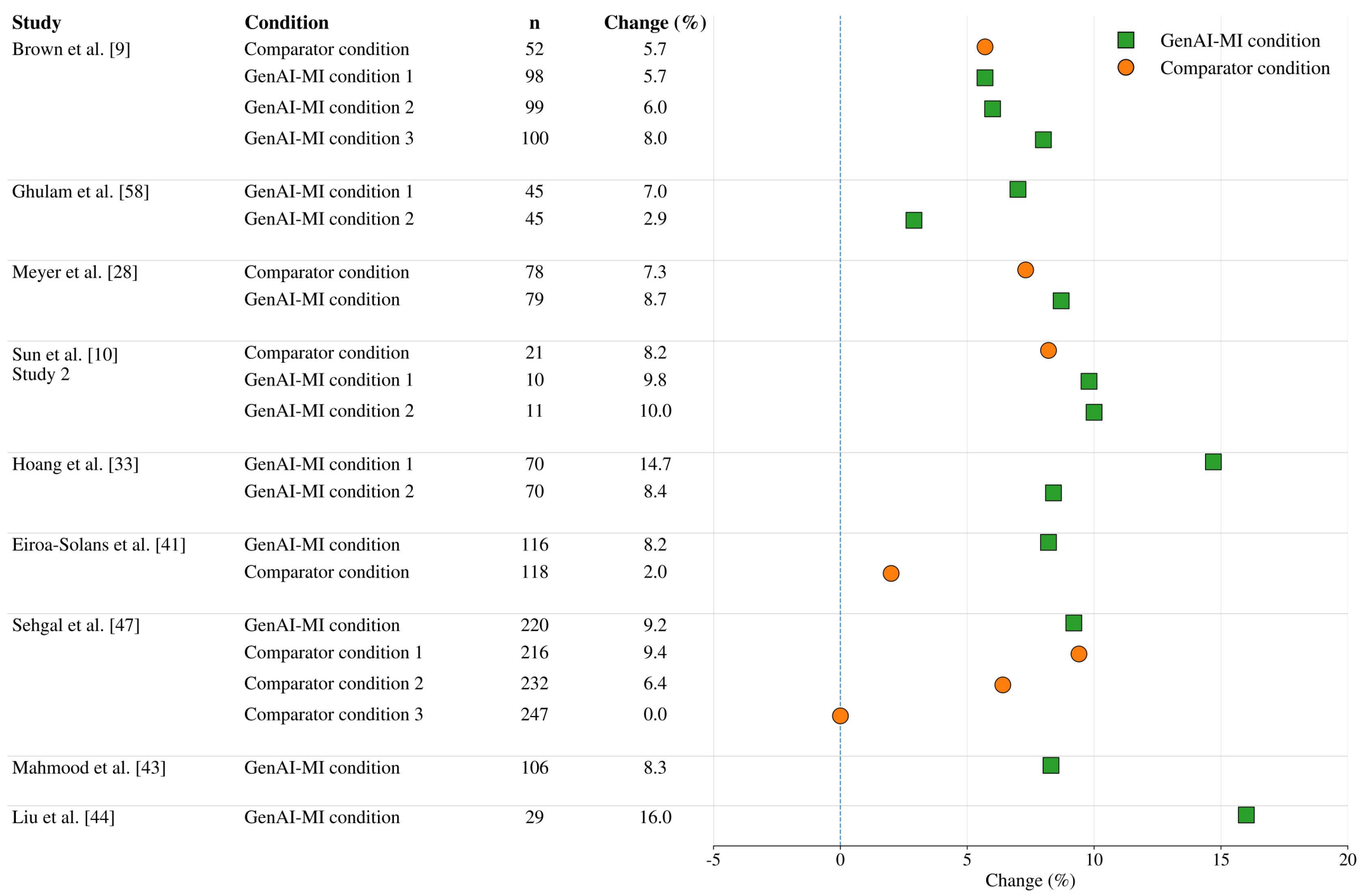

Note: range-normalized changes in motivation outcomes were calculated as the pre-post mean change divided by the full scale range (%). Only studies for which this value could be calculated were included. Orange circles indicate comparator conditions, and green squares indicate GenAI-MI conditions.

# Discussion

## Principal Findings

This scoping review systematically synthesized the current evidence on GenAI-MI chatbots across five domains: system design, safety measures, MI quality evaluation, user perceptions, and intervention outcomes, encompassing 48 studies in total. This scoping review shows that research on GenAI-MI chatbots has progressed rapidly from system development toward direct use in health and behavior change contexts, but the evidence remains concentrated at relatively early stages of intervention translation. The current literature suggests that although the field has progressed further in demonstrating the delivery of MI through GenAI, considerable further evidence is needed to establish its effectiveness in producing meaningful and sustained behavior change.

Across this evidence continuum, several patterns emerged. GenAI-MI systems remained predominantly text based and disembodied, although dynamic adaptation was already common. Safety practices relied more heavily on preventive safeguards than on runtime monitoring. MI-specific observer and client evaluations generally suggested favorable MI quality, whereas generic natural language generation metrics provided limited information about MI fidelity. Participants also generally reported favorable perceptions of the systems. However, intervention evidence was dominated by single-session studies and proximal motivation outcomes, with substantially less evidence concerning repeated use, sustained behavior change, or functional outcomes.

## Interpretation of Findings

Regarding system design, GenAI-MI chatbots remained predominantly text based and disembodied, suggesting that most current systems continue to rely on relatively conventional conversational interfaces despite the broader technical possibilities enabled by GenAI. At the same time, dynamic adaptation was incorporated in nearly half of the included studies, using information such as inferred motivational states, selected MI strategies and longitudinal user information. This development is relevant to MI because effective MI requires ongoing responsiveness to clients’ expressions, motivational states, ambivalence, and readiness for change, with conversational strategies adjusted as the interaction unfolds[62]. Speech, visual interfaces, virtual agents, and physical robots further expanded how MI could be delivered, but these approaches remained relatively uncommon.

The safety findings showed that privacy and data protection, together with safety-oriented content generation, were more commonly described, whereas runtime automated risk monitoring and human risk monitoring were reported in relatively few studies. This pattern suggests that current safety practices have focused more on preventive safeguards than on detecting and responding to risks that emerge during ongoing interactions. A previous review focusing on the safety of GenAI mental health chatbots suggested that prompts or automated monitoring alone are unlikely to address all risks arising from open generation, and that more comprehensive safeguards require both human oversight and clearly defined risk escalation pathways[14]. Among studies involving direct participant use, informed consent or user education was commonly reported, reflecting substantial attention to participant safety and research ethics.

For MI quality, the findings highlight an important distinction between general language generation quality and fidelity to MI. Natural language generation metrics primarily capture lexical or semantic similarity between generated responses and reference texts and therefore cannot adequately assess the quality of MI delivery itself. In some studies, their results were inconsistent with evaluations by MI experts. In contrast, studies using MI quality assessment methods such as MITI, MISC, and CEMI generally indicated that existing GenAI-MI chatbots were capable of delivering relatively high-quality MI. A previous systematic review of MI quality assessment tools likewise emphasized the importance of using dedicated quality measures to determine whether MI is delivered appropriately[15]. Therefore, natural language generation metrics alone are insufficient for evaluating the MI quality of GenAI-MI chatbots. Some studies have also begun to use LLMs or dedicated classification models for automated coding of MI dialogue, achieving moderate to good agreement with human coding and potentially making systematic monitoring of MI process quality more feasible in large scale digital health interventions. However, automated coding may still be prone to errors when assessing complex and context-dependent MI behaviors. Continued validation against human expert coding within standardized coding frameworks is therefore needed[63, 64].

User perceptions were generally favorable, particularly in terms of empathy, usability, helpfulness, and intention to use. These findings suggest that users may find GenAI-MI interactions acceptable and relationally engaging, which may facilitate their use in behavioral interventions. However, the measurement of user perceptions was highly heterogeneous, limiting direct comparisons across studies. Similar heterogeneity in how users perceive health care chatbots has been reported, highlighting the need to assess user perceptions across multiple dimensions[65].

For intervention outcomes, favorable findings were reported more consistently for proximal motivation outcomes, including readiness to change, confidence, and intention, than for behavioral or functional outcomes. However, these findings were derived from heterogeneous study designs, many involving single-session or uncontrolled evaluations. Accordingly, the current evidence suggests a generally favorable pattern for short-term motivational outcomes but does not establish the effectiveness of GenAI-MI chatbots as behavioral interventions. Behavioral and functional outcomes were reported less frequently, and evidence for sustained or objectively measured behavioral change remained limited and mixed.

## Future Research

Future research should further evaluate the practical value of the system design features identified in this review. Speech, visual interfaces, virtual or physical embodiment, and dynamic adaptation may broaden how GenAI-MI chatbots are delivered and tailored, but additional complexity should not be assumed to improve performance. Direct comparative studies should determine whether dynamic adaptation and multimodal interaction meaningfully improve MI quality, user perceptions, or intervention outcomes relative to simpler designs. As GenAI-MI chatbots are increasingly used in complex and sensitive health settings, stronger safety safeguards are needed. Future systems should include runtime risk monitoring and clear pathways for human intervention. This is particularly important in applications involving substance use, psychological support, or health advice. Future studies should also define how high-risk content is identified and escalated, while ensuring appropriate informed consent and data protection. Evaluation methods for GenAI-MI chatbots also require further standardization. For systems in which MI is a core component, MI quality should be a key evaluation target, and evaluations should not rely primarily on natural language generation metrics. In studies involving direct participant use of chatbots, client MI measures such as CEMI may be used to complement observer coding. Automated MI coding also offers new possibilities for dialogue assessment at scale. Future research should further examine its accuracy and stability and promote its use in GenAI-MI chatbot evaluation only after sufficient validation against human coding. More importantly, future studies should strengthen the evaluation of sustained use and behavioral outcomes. Most existing intervention studies involve only a single session or short-term interactions, and their outcomes are concentrated on short-term motivational indicators. Subsequent studies should extend intervention and follow-up periods, include appropriate comparator conditions, and incorporate behavioral measures that are directly related to the intervention target. Such studies are needed to determine whether the short-term positive changes in motivation outcomes reported following GenAI-MI chatbot use can be sustained and ultimately translated into actual behavioral change.

## Strengths and Limitations

This review has several strengths. First, to our knowledge, no previous scoping review has specifically focused on GenAI-MI chatbots. Building on the existing literature, this review provides an integrated synthesis of system design, safety measures, MI quality evaluation, user perceptions, and intervention outcomes, thereby offering a relatively comprehensive overview of this emerging field. Second, the review included studies spanning the design, feasibility, and evaluation study types, enabling the field to be examined across the continuum from technological development to practical application. In addition, we searched multiple databases across medicine, psychology, and computer science. We also included conference proceedings and preprints, which helped capture rapidly emerging evidence in this evolving research area.

This review also has several limitations. First, only publications in English were included, which may have resulted in the omission of relevant studies published in other languages. Second, the included studies were highly heterogeneous, limiting evidence synthesis and the generalizability of the findings. We therefore primarily used descriptive statistics and narrative synthesis rather than pooling effect sizes. The standardized presentation of motivational changes was intended to facilitate visualization of the distribution of findings across studies and should not be interpreted as supporting inferential comparisons of intervention effects between studies. Third, we did not conduct a formal risk of bias or critical appraisal of the included studies. Therefore, the reported intervention outcomes should be interpreted as a descriptive characterization of the existing literature rather than as evidence of intervention effectiveness or certainty of effect.

## Conclusions

Overall, existing studies have generally reported positive user perceptions and short-term increases in motivation outcomes. Most systems remain text-based and disembodied, while dynamic adaptation has already been incorporated in nearly half of the included studies and a smaller number have explored multimodal and embodied interaction. However, current research remains concentrated on short-term interactions and proximal outcomes, while sustained use and behavioral or functional outcomes have been examined less frequently. Alongside continued innovation in GenAI-MI chatbot systems, future research should strengthen safety measures and MI quality evaluation and conduct longer-term interventions with assessment of real-world behavioral outcomes, thereby providing a clearer understanding of the practical value of GenAI-MI chatbots in behavior change applications.

## Data availability

This scoping review synthesizes data extracted from previously published studies, and the extracted and synthesized data supporting the findings are available (Multimedia Appendix 4). Data from the included studies are available from the original publications or from the respective authors according to their data availability statements.

## Funding

This study was supported by National Natural Science Foundation of China (82373694) and Beijing Natural Science Foundation (7262165). The funders had no role in the design of the review, study selection, data extraction, data synthesis, interpretation of the findings, or manuscript preparation.

## Conflicts of Interest

None declared.

## Authors' Contributions

R.H. and J.K. conducted the literature screening, data extraction, and analysis. R.H. wrote the original draft of the manuscript. Y.Y., Y.H.Y., J.L., H.T., and S.Z. checked and verified the results. Z.L. supervised the analysis and manuscript preparation and critically revised the manuscript. All authors reviewed and approved the final manuscript.

## Acknowledgments

In accordance with the GAIDeT taxonomy (2025), the authors declare the use of ChatGPT-5.6 under full human supervision for literature summarization, visualization, proofreading, editing, and translation during the research and manuscript preparation. Responsibility for the final manuscript lies entirely with the authors. GAI tools are not listed as authors.

## Abbreviations

CEMI: Client Evaluation of Motivational Interviewing

GenAI: generative artificial intelligence

LLM: large language model

MI: motivational interviewing

MISC: Motivational Interviewing Skill Code

MITI: Motivational Interviewing Treatment Integrity

## References

1. Naslund JA, Aschbrenner KA, Araya R, Marsch LA, Unützer J, Patel V, et al. Digital technology for treating and preventing mental disorders in low-income and middle-income countries: a narrative review of the literature. Lancet Psychiatry. 2017 Jun;4(6):486–500. PMID: 28433615. doi: 10.1016/s2215-0366(17)30096-2.
2. Tudor Car L, Dhinagaran DA, Kyaw BM, Kowatsch T, Joty S, Theng YL, et al. Conversational Agents in Health Care: Scoping Review and Conceptual Analysis. J Med Internet Res. 2020 Aug 7;22(8):e17158. PMID: 32763886. doi: 10.2196/17158.
3. Hua Y, Na H, Li Z, Liu F, Fang X, Clifton D, et al. A scoping review of large language models for generative tasks in mental health care. npj Digital Medicine. 2025 04/30;8(1):230. doi: 10.1038/s41746-025-01611-4.
4. Chen J, Hu R-Z, Zhuang Y-X, Zhang J-Q, Shan R, Yang Y, et al. Natural Language Processing Chatbot–Based Interventions for Improvement of Diet, Physical Activity, and Tobacco Smoking Behaviors: Systematic Review. JMIR Mhealth Uhealth. 2025;13:e66403. doi: 10.2196/66403.
5. Miller WR, Rollnick S. Motivational Interviewing: Helping People Change and Grow. 4th ed. New York, NY: The Guilford Press; 2023. ISBN: 9781462552795.
6. Moyers TB, Manuel JK, Ernst D. Motivational Interviewing Treatment Integrity Coding Manual 4.2.1. Albuquerque: Center on Alcoholism, Substance Abuse and Addictions, University of New Mexico; 2015.
7. Houck JM, Moyers TB, Miller WR, Glynn LH, Hallgren KA. Motivational Interviewing Skill Code (MISC) 2.5. Albuquerque: Center on Alcoholism, Substance Abuse and Addictions, University of New Mexico; 2010.
8. Yang Y, Achananuparp P, Huang H, Jiang J, Kit PL, Lim NG, et al. CAMI: A Counselor Agent Supporting Motivational Interviewing through State Inference and Topic Exploration. July; Vienna, Austria: Association for Computational Linguistics; 2025. p. 21037–81. doi: 10.18653/v1/2025.acl-long.1024.
9. Brown A, Kumar AT, Melamed O, Ahmed I, Wang YH, Deza A, et al. A motivational interviewing chatbot with generative reflections for increasing readiness to quit smoking: Iterative development study. JMIR Mental Health. 2023;10. doi: 10.2196/49132.
10. Sun X, Wit Jd, Li Z, Pei J, Ali AE, Bosch JA. Script-Strategy Aligned Generation: Aligning LLMs with Expert-Crafted Dialogue Scripts and Therapeutic Strategies for Psychotherapy. Proc ACM Hum-Comput Interact. 2025;9(7):Article CSCW474. doi: 10.1145/3757655.
11. Karve Z, Calpey J, Machado C, Knecht M, Mejia MC. New Doc on the Block: Scoping Review of AI Systems Delivering Motivational Interviewing for Health Behavior Change. J Med Internet Res. 2025;27:e78417. doi: 10.2196/78417.
12. Tricco AC, Lillie E, Zarin W, O'Brien KK, Colquhoun H, Levac D, et al. PRISMA Extension for Scoping Reviews (PRISMA-ScR): Checklist and Explanation. Annals of Internal Medicine. 2018;169(7):467–73. PMID: 30178033. doi: 10.7326/m18-0850.
13. Balan R, Dobrean A, Poetar CR. Use of automated conversational agents in improving young population mental health: a scoping review. npj Digital Medicine. 2024 03/19;7(1):75. doi: 10.1038/s41746-024-01072-1.
14. Olisaeloka L, Richardson CG, Wang AY, Munthali RJ, Vigo DV. Safety Mechanisms and Risk Mitigation in Generative AI Mental Health Chatbots: A Systematic Scoping Review. Healthcare. 2026;14(10):1395. doi: 10.3390/healthcare14101395.
15. Hurlocker MC, Madson MB, Schumacher JA. Motivational interviewing quality assurance: A systematic review of assessment tools across research contexts. Clinical Psychology Review. 2020 12/01/;82:101909. doi: 10.1016/j.cpr.2020.101909.

16. Li H, Zhang R, Lee Y-C, Kraut RE, Mohr DC. Systematic review and meta-analysis of AI-based conversational agents for promoting mental health and well-being. npj Digital Medicine. 2023 12/19;6(1):236. doi: 10.1038/s41746-023-00979-5.
17. López-López JA, Page MJ, Lipsey MW, Higgins JPT. Dealing with effect size multiplicity in systematic reviews and meta-analyses. Research Synthesis Methods. 2018 09/01;9(3):336–51. doi: 10.1002/jrsm.1310.
18. Nie J, Shao H, Fan Y, Shao Q, You H, Preindl M, et al. LLM-based Conversational AI Therapist for Daily Functioning Screening and Psychotherapeutic Intervention via Everyday Smart Devices. ACM Transactions on Computing for Healthcare. 2026;7(3). doi: 10.1145/3712299.
19. Jörke M, Genç D, Teutschbein V, Sapkota S, Chung S, Schmiedmayer P, et al. Bloom: Designing for LLM-Augmented Behavior Change Interactions. Proceedings of the 2026 CHI Conference on Human Factors in Computing Systems: Association for Computing Machinery; 2026. p. Article 1167. doi: 10.1145/3772318.3790506.
20. Jörke M, Sapkota S, Warkenthien L, Vainio N, Schmiedmayer P, Brunskill E, et al. GPTCoach: Towards LLM-Based Physical Activity Coaching. PROCEEDINGS OF THE 2025 CHI CONFERENCE ON HUMAN FACTORS IN COMPUTING SYSTEMS, CHI 2025. doi: 10.1145/3706598.3713819.
21. Younsi N, Pelachaud C, Chaby L. MODIFF-8 to better motivate: Live adaptive human-socially interactive agent interaction. JOURNAL ON MULTIMODAL USER INTERFACES. 2026 MAR 25. doi: 10.1007/s12193-026-00477-4.
22. Ashraf R, Liu Y, Ganji S, Kim JH. "I Loved How Pepper Talked About My Hoodie": A Situated, MI-Grounded Multimodal Architecture for Engaging Conversation. Companion Proceedings of the 21st ACM/IEEE International Conference on Human-Robot Interaction; Edinburgh, Scotland, UK: Association for Computing Machinery; 2026. p. 639–43. doi: 10.1145/3776734.3794474.
23. Yeo YH, Clark A, Mehra M, Danovitch I, Osilla K, Yang JD, et al. The Feasibility and Usability of an Artificial Intelligence-Enabled Conversational Agent in Virtual Reality for Patients with Alcohol-Associated Cirrhosis: A Multi-Methods Study. Journal of Medical Extended Reality. 2024;1(1). doi: 10.1089/jmxr.2024.0033.
24. Olafsson S, Pedrelli P, Wallace BC, Bickmore T. Accomodating User Expressivity while Maintaining Safety for a Virtual Alcohol Misuse Counselor. Proceedings of the 23rd ACM International Conference on Intelligent Virtual Agents; Würzburg, Germany: Association for Computing Machinery; 2023. p. Article 3. doi: 10.1145/3570945.3607361.
25. Galland L, Pelachaud C, Pecune F. SMART-DREAM: To Condition or Not to Condition; A Study on the Impact of LLM Conditioning on Motivational Interview Dialog Virtual Agent. Proceedings of the 25th ACM International Conference on Intelligent Virtual Agents: Association for Computing Machinery; 2025. p. Article 4. doi: 10.1145/3717511.3747062.
26. Steenstra I, Nouraei F, Arjmand M, Bickmore TW. Virtual Agents for Alcohol Use Counseling: Exploring LLM-Powered Motivational Interviewing. PROCEEDINGS OF THE 24TH ACM INTERNATIONAL CONFERENCE ON INTELLIGENT VIRTUAL AGENTS, IVA 2024. doi: 10.1145/3652988.3673932.
27. Galland L, Pelachaud C, Pecune F. Tailored Conversations beyond LLMs: A RL-Based Dialogue Manager. arXiv. arXiv2025.
28. Meyer S, Elsweiler D. LLM-based conversational agents for behaviour change support: A randomised controlled trial examining efficacy, safety, and the role of user behaviour. INTERNATIONAL JOURNAL OF HUMAN-COMPUTER STUDIES. 2025 MAY;200. doi: 10.1016/j.ijhcs.2025.103514.

29. Hu R-z, Yang Y, Yang Y-h, Kong J-q, Luo J-h, Yang W-y, et al. Fine-Tuning Large Language Models for Motivational Interviewing in Health Behavior Change: Development and Evaluation Study. JMIR Formative Research. 2026;10:e89077. doi: 10.2196/89077.

30. Kim H, Lee S, Cho Y, Ryu E, Jo Y, Seong S, et al. KMI: A Dataset of Korean Motivational Interviewing Dialogues for Psychotherapy. April; Albuquerque, New Mexico: Association for Computational Linguistics; 2025. p. 10803–28. doi: 10.18653/v1/2025.naacl-long.541.

31. Zeng J, Nakano YI. Schema-Guided Response Generation using Multi-Frame Dialogue State for Motivational Interviewing Systems. Findings of the Association for Computational Linguistics: ACL 2026: Association for Computational Linguistics; 2026. p. 41493–524. doi: 10.18653/v1/2026.findings-acl.2063.

32. Tao J, Pavlick E, Grondin A, Bustamante JD, Martin H, Parent H, et al. Evaluation of an Artificial Intelligence Conversational Chatbot to Enhance HIV Preexposure Prophylaxis Uptake: Development and Usability Internal Testing. J Med Internet Res. 2026 Feb 3;28:e79671. PMID: 41632955. doi: 10.2196/79671.

33. Hoang V, Rogers E, Ross RJ. An LLM-Based Motivation-Aware Framework For AI Coaching For Behaviour Change. Proceedings of the 2026 CHI Conference on Human Factors in Computing Systems: Association for Computing Machinery; 2026. p. Article 1166. doi: 10.1145/3772318.3791123.

34. Sun X, Tang X, El Ali A, Li Z, Ren P, de Wit J, et al. Rethinking the Alignment of Psychotherapy Dialogue Generation with Motivational Interviewing Strategies. Proceedings of the 31st International Conference on Computational Linguistics. 2025.

35. Wang L CD, Filienko D, El Jazmi C, Xie SJ, De Cock M, Iribarren S, Yuwen W. Large Language Model-Powered Conversational Agent Delivering Problem-Solving Therapy (PST) for Family Caregivers: Enhancing Empathy and Therapeutic Alliance Using In-Context Learning. AMIA Annual Symposium Proceedings 2025. p. 1315–24.

36. Xi ZH, Majumder BP, Zhao MJ, Maeda Y, Yamada K, Wakaki H, et al. Few-shot Dialogue Strategy Learning for Motivational Interviewing via Inductive Reasoning. FINDINGS OF THE ASSOCIATION FOR COMPUTATIONAL LINGUISTICS: ACL 2024. p. 13207–19. doi: 10.18653/v1/2024.findings-acl.782.

37. Wang J, Yao Z, Li L, Qian J, Yang Z, Yu H. ChatThero: An LLM-Supported Chatbot for Behavior Change and Therapeutic Support in Addiction Recovery. 2025. 10.48550/arXiv.2508.20996

38. Sun X, Krahmer E, Wit JD, Wiers R, Bosch JA. Plug and Play Conversations: The Micro-Conversation Scheme for Modular Development of Hybrid Conversational Agent. Companion Publication of the 2023 Conference on Computer Supported Cooperative Work and Social Computing; Minneapolis, MN, USA: Association for Computing Machinery; 2023. p. 50–5. doi: 10.1145/3584931.3606998.

39. Teferra BG, Huang S, Johny N, Perivolaris A, Al-Shamali H, Parkington K, et al. Alignment of Large Language Model Responses With Human Therapists in Motivational Interviewing. JAMA Netw Open. 2026 Mar 2;9(3):e262750. PMID: 41870428. doi: 10.1001/jamanetworkopen.2026.2750.

40. Jha A, Shivaprakash P, Shukla L, Mukherjee A, Chand P, Murthy P. Benchmarking Motivational Interviewing Competence of Large Language Models. European Addiction Research. 2026:1–18. doi: 10.1159/000553455.

41. Eiroa-Solans C, Inzlicht M. From extrinsic to intrinsic motivation: Testing an AI-powered motivational interviewing system to foster prosocial motivation. COMPUTERS IN HUMAN BEHAVIOR REPORTS. 2026 MAR;21. doi: 10.1016/j.chbr.2025.100882.

42. Suffoletto B, Prieksaitis C, Rose C, Kim D, Pillai N, Pitre V. Feasibility and Acceptability of a Large Language Model-Based Motivational Interviewing Agent in the Emergency Department. Annals of Emergency Medicine. 2026 01/01;87(1):121–3. doi: 10.1016/j.annemergmed.2025.08.021.
43. Mahmood Z, Ali S, Zhu J, Abdelwahab M, Collins MY, Chen S, et al. A Fully Generative Motivational Interviewing Counsellor Chatbot for Moving Smokers Towards the Decision to Quit. Findings of the Association for Computational Linguistics: ACL 2025. p. 25008–43. doi: 10.18653/v1/2025.findings-acl.1283.
44. Liu Y, Calle P, Vadakekut M, Rubin D, Nagykaldi Z, Doescher M, et al. AI-Enabled Personalized Smoking Cessation Intervention With the Aipaca Chatbot: Mixed Methods Feasibility Study. JMIR Form Res. 2025 Dec 11;9:e73319. PMID: 41380150. doi: 10.2196/73319.
45. Bak M, Quan K, Tomaszewski T, Chin J. Can Conversational AI Counsel for Change? A Theory-Driven Approach to Supporting Dietary Intentions in Ambivalent Individuals. 2025. 10.48550/arXiv.2511.02428
46. Suffoletto B, Clark DB, Lee C, Mason M, Schultz J, Szeto I, et al. Development and preliminary testing of a secure large language model-based chatbot for brief alcohol counseling in young adults. Drug and Alcohol Dependence. 2025;272:1–5. doi: 10.1016/j.drugalcdep.2025.112697.
47. Sehgal NKR, Tonneau M, Tan A, Mehta SJ, Buttenheim A, Ungar L, et al. Effect of Static vs. Conversational AI-Generated Messages on Colorectal Cancer Screening Intent: a Randomized Controlled Trial. 2025. 10.48550/arXiv.2507.08211
48. Herbert D, Westendorf J, Farmer M, Reeder B. Generative AI-derived information about opioid use disorder treatment during pregnancy: An exploratory evaluation of GPT-4's steerability for provision of trustworthy person-centered information. Journal of Studies on Alcohol and Drugs. 2025;86(6):894–905. doi: 10.15288/jsad.24-00319.
49. Bolpagni M, De Carli S, Sanna L, Gabrielli S, Dragoni M. Role-Play Large Language Models for Short Behavior Change Interventions: An Exploratory Study on Brief Action Planning. ARTIFICIAL INTELLIGENCE IN MEDICINE, AIME 2025, PT II2025. p. 46–51. doi: 10.1007/978-3-031-95841-0_9.
50. Yosef S, Zisquit M, Cohen B, Klomek AB, Bar K, Friedman D. The impact of fine-tuning LLMs on the quality of automated therapy assessed by digital patients. Npj Ment Health Res. 2025 Sep 13;4(1):43. PMID: 40946097. doi: 10.1038/s44184-025-00159-1.
51. Huang Y, Jiang Y, Liu H, Cai Y, Li W, Hu X. AI-Augmented LLMs Achieve Therapist-Level Responses in Motivational Interviewing. 2025. 10.48550/arXiv.2505.17380
52. Meywirth S. Designing a Large Language Model-Based Coaching Intervention for Lifestyle Behavior Change. DESIGN SCIENCE RESEARCH FOR A RESILIENT FUTURE, DESRIST 2024. p. 81–94. doi: 10.1007/978-3-031-61175-9_6.
53. Bilancini E, Boncinelli L, Vicario E. AI-powered Chatbots: Effective Communication Styles for Sustainable Development Goals. 2024. 10.48550/arXiv.2407.01057
54. Gabriel S, Puri I, Xu XH, Malgaroli M, Ghassemi M. Can AI Relate: Testing Large Language Model Response for Mental Health Support. FINDINGS OF THE ASSOCIATION FOR COMPUTATIONAL LINGUISTICS: EMNLP 2024. p. 2206–21. doi: 10.18653/v1/2024.findings-emnlp.120.
55. Kumar AT, Wang C, Dong A, Rose J. Generation of Backward-Looking Complex Reflections for a Motivational Interviewing-Based Smoking Cessation Chatbot Using GPT-4: Algorithm Development and Validation. JMIR Ment Health. 2024 Sep 26;11:e53778. PMID: 39324852. doi: 10.2196/53778.
56. Basar E, Hendrickx I, Krahmer E, Bruijn G-J, Bosse T. To What Extent Are Large Language Models Capable of Generating Substantial Reflections for Motivational Interviewing Counseling

Chatbots? A Human Evaluation. Proceedings of the 1st Human-Centered Large Language Modeling Workshop. 2024. doi: 10.18653/v1/2024.hucllm-1.4.

57. Brown A, Zhu JD, Abdelwahab M, Dong A, Wang C, Rose J. Generation, Distillation and Evaluation of Motivational Interviewing-Style Reflections with a Foundational Language Model. PROCEEDINGS OF THE 18TH CONFERENCE OF THE EUROPEAN CHAPTER OF THE ASSOCIATION FOR COMPUTATIONAL LINGUISTICS, VOL 1: LONG PAPERS2024. p. 1241–52. doi: 10.18653/v1/2024.eacl-long.75.

58. Ghulam H, Keegan B, Ross R. Active Listening in Virtual Interactive Coaching: Prompt Strategies and User Assessment. Proceedings of the 38th International BCS Human-Computer Interaction Conference; Cardiff University, UK: BCS Learning & Development Ltd; 2025. p. 550–64. doi: 10.14236/ewic/BCSHCI2025.61.

59. Zhu JD, Dong A, Wang C, Veldhuizen S, Abdelwahab M, Brown A, et al. The Impact of ChatGPT Exposure on User Interactions With a Motivational Interviewing Chatbot: Quasi-Experimental Study. JMIR FORMATIVE RESEARCH. 2025;9. doi: 10.2196/56973.

60. Meng Q, Chen M, Liu D, Mo Y, Su Y, Sun X, et al. StoryMI: Steerable Multi-Agent Therapeutic Dialogue Generation. Findings of the Association for Computational Linguistics: ACL 2026. p. 9606–23. doi: 10.18653/v1/2026.findings-acl.468.

61. Kim J, Rodriguez VJ, Yoo DW, Chandrasekharan E, Saha K. PAIR-SAFE: A Paired-Agent Approach for Runtime Auditing and Refining AI-Mediated Mental Health Support. 2026. 10.48550/arXiv.2601.12754

62. Miller WR, Rose GS. Toward a theory of motivational interviewing. American Psychologist. 2009;64(6):527–37. doi: 10.1037/a0016830.

63. Ali S, Zhu J, Guo A, Ye XN, Gu Q, Wolff J, et al. Automated Coding of Counsellor and Client Behaviours in Motivational Interviewing Transcripts: Validation and Application. NLP-AI4Health. 2025. doi: 10.18653/v1/2025.nlpai4health-main.4.

64. Pellemans M, Salmi S, Mérelle S, Janssen W, van der Mei R. Automated behavioral coding to enhance the effectiveness of motivational interviewing in a chat-based suicide prevention helpline: Secondary analysis of a clinical trial. Journal of Medical Internet Research. 2024;26. doi: 10.2196/53562.

65. Hua Y, Xia W, Bates D, Hartstein GL, Kim HT, Li M, et al. Standardizing and Scaffolding Health Care AI-Chatbot Evaluation: Systematic Review. JMIR AI. 2025;4:e69006. doi: 10.2196/69006.